\documentclass[review,authoryear]{elsarticle}
\usepackage{array}
\newcolumntype{P}[1]{>{\centering\arraybackslash}p{#1}}
\usepackage{lineno,hyperref}
\usepackage{graphicx}  
\usepackage[labelsep=quad,indention=10pt]{subfig}
\usepackage{caption}
\usepackage{ dsfont }
\usepackage{float}
\usepackage{soul}
\usepackage{multirow}
\usepackage[dvipsnames]{xcolor}

\usepackage{todonotes}

\makeatletter
\def\fixedlabel#1#2{%
  \@bsphack%
  \protected@write\@auxout{}%
         {\string\newlabel{#1}{{#2}{\thepage}}}%
  \@esphack}
\makeatother

\journal{Computers and Electronics in Agriculture}

\begin{document}

\begin{frontmatter}

\title{Semantic Segmentation for Partially Occluded Apple Trees Based on Deep Learning}

\author{Zijue Chen}
\author{David Ting}
\author{Rhys Newbury}
\author{Chao Chen\corref{mycorrespondingauthor}}

\cortext[mycorrespondingauthor]{Corresponding author}
\ead{chao.chen@monash.edu}

\address{Laboratory of Motion Generation and Analysis, Faculty of Engineering, Monash University, Clayton, VIC 3800,Australia}

\begin{abstract}

Fruit tree pruning and fruit thinning require a powerful vision system that can provide high resolution segmentation of the fruit trees and their branches. However, recent works only consider the dormant season, where there are minimal occlusions on the branches or fit a polynomial curve to reconstruct branch shape and hence, losing information about branch thickness. In this work, we apply two state-of-the-art supervised learning models U-Net and DeepLabv3, and a conditional Generative Adversarial Network Pix2Pix (with and without the discriminator) to segment partially occluded 2D-open-V apple trees. Binary accuracy, Mean IoU, Boundary F1 score and Occluded branch recall were used to evaluate the performances of the models. DeepLabv3 outperforms the other models at Binary accuracy, Mean IoU and Boundary F1 score, but is surpassed by Pix2Pix (without discriminator) and U-Net in Occluded branch recall. We define two difficulty indices to quantify the difficulty of the task: (1) Occlusion Difficulty Index and (2) Depth Difficulty Index. We analyze the worst 10 images in both difficulty indices by means of \textit{Branch} Recall and \textit{Occluded Branch} Recall. U-Net outperforms the other two models in the current metrics. On the other hand, Pix2Pix (without discriminator) provides more information on branch paths, which are not reflected by the metrics. This highlights the need for more specific metrics on recovering occluded information. Furthermore, this shows the usefulness of image-transfer networks for hallucination behind occlusions. Future work is required to further enhance the models to recover more information from occlusions such that this technology can be applied to automating agricultural tasks in a commercial environment.

\end{abstract}


\end{frontmatter}


\section{Introduction}

In the last few years there has been a growing demand for automation in the agricultural industry. Within this industry, apple farming is one of the largest crop harvest and production sectors. Increasing labor shortages and costs have resulted in a larger focus on automation of harvesting and maintenance of the apple orchards \citep{apple_review}.

 Apart from harvesting, there are two major tasks to the maintenance of an apple farm, fruit thinning and tree pruning. Apple tree pruning occurs in both the winter and summer seasons to trim obstructing branches to encourage better growth and flowering from the trees. Apple thinning is the process of removing damaged, small or clustered apples to improve the apple size and quality of the remaining apples. An important factor that determines the optimal amount of apples on an apple branch is the branch thickness. Therefore, in order to effectively perform automated apple thinning, the thickness along the branch will need to be determined as well as which apples are attached to that branch. Both of these require an accurate structural understanding of the whole apple tree. Therefore, algorithms need to be developed which can successfully detect and reconstruct branches accurately. In this work, the branch reconstruction focuses only on reconstructing the main branches off the trunk where apples are likely to grow during the thinning season.

 To detect the branches of the tree, a branch of deep learning, semantic segmentation is often applied. Semantic Segmentation is a well studied problem in computer vision~\citep{ulku2019survey}. Many different network structures have been proposed with high accuracy in pixel wise masking. Mask R-CNN~\citep{maskrcnn} extends Faster R-CNN to segment both the bounding boxes and the segmented image simultaneously~\citep{ren2015faster}. This was applied by \citet{ZHANG2018386} to successfully segment branches and trunks. \citet{MAJEED201875} utilized SegNet~\citep{segnet}, a fully convolutional neural network architecture, to segment branches and trunks. U-Net \citep{unet} focuses on data efficiency to train a convolutional neural network by utilizing feature map skip connections and was  utilized to segment visible litchi stem at a $150$*$150$ Region of Interest \citep{LIANG2020105192}. \citet{KANG2020105302} created a lightweight network, DaSNet-v2, to segment both apples and branches simultaneously.

 However, these works either only consider visible branches during dormant seasons, where there were minimal occlusions, or only fit a polynomial curves to reconstruct branch shape, providing a lack of information on branch thickness. 
 
 \citet{purkait2019seeing} investigated using semantic segmentation for occluded objects utilizing the U-Net~\citep{unet} segmentation model with a custom loss function. Their results conveyed that supervised learning techniques were capable of performing occluded segmentation. 
 
 In order to create a system capable of tasks such as fruit thinning and pruning, the method must be robust from the expected increasing amounts of occlusion in the seasons those tasks take place. Thus, a robust vision system that can accurately identify branch systems under occlusion is required before these tasks can be automated in a commercial environment. 
 
DeepLabv3~\citep{deeplab} is a current state-of-the-art model in the semantic segmentation domain, scoring the highest overall accuracy based on the VOC2012 dataset~\citep{pascalVoc}. U-Net and DeepLabv3 will be explored as the current representations of the best segmentation network for visible panoptic segmentation.

Apart from Convolutional Neural Network architectures, Generative Adversarial Networks (GANs) have also been implemented into semantic pixel-wise segmentation. By adding in a discriminator, higher-order patterns and relationships can be learned which can improve the results' accuracy on the original segmentation networks~\citep{Luc2016}. Although most GANs are used to generate realistic images, conditional GANs with additional input can be used for special processing of input images. The Pix2Pix model~\citep{pix2pix} is a conditional GAN which has shown good performance in image transfer problems and has been previously applied to semantic segmentation successfully~\citep{pix2pixsegment}.
 
This paper explores methodologies and techniques in order to improve evaluation and accuracy of occluded apple tree segmentation. Specifically, this paper aims to address the research gap of segmentation of branches under occlusion of leaves.

The contributions of this paper are three-fold:
\begin{enumerate}
    \item Successful application of state-of-the-art segmentation models to occluded branches with a comprehensive comparison.
    \item Successful application of Generative Adversarial architecture to segmentation of occluded branches with the emphasis of unsupervised learning.
    \item Development of effective indices for occluded oriented evaluation.
\end{enumerate}

The organization of this work is as follows. The detailed methodology is described in Section~\ref{sec:Methodology}. The results and discussion are shown in Section~\ref{sec:r+d}. Finally the conclusion is presented in Section~\ref{sec:conclusion}.

\section{Methodology}
\label{sec:Methodology}

\subsection{Dataset Collection}

The dataset images used in this paper were taken from a commercial apple orchard in northeastern Melbourne. The apple species is Ruby Pink and the apple tree structure is 2D-open-V (Tatura Trellis). There is $4.125$ m between the centers of each row of trees with $0.6$ m between each tree in each row, planted east to west. The distance between camera and apple tree canopies is between $1.8$ m to $2.2$ m. The pictures were taken at the end of May, just after the apple harvest season in Melbourne. As a result, there are no occlusions due to apples while the majority of leaves are still attached to the tree. The data was collected on a clear weather sunny and cloudless autumn afternoon between 4 pm to 5.30 pm, before a sunset time of 6.15 pm. The Intel Realsense D435 was utilized to take RGB and depth photos in a $640$ by $480$ pixel resolution and aligning the depth images to the RGB images.

\begin{figure}[h!]
\centering
\includegraphics[width=1\textwidth]{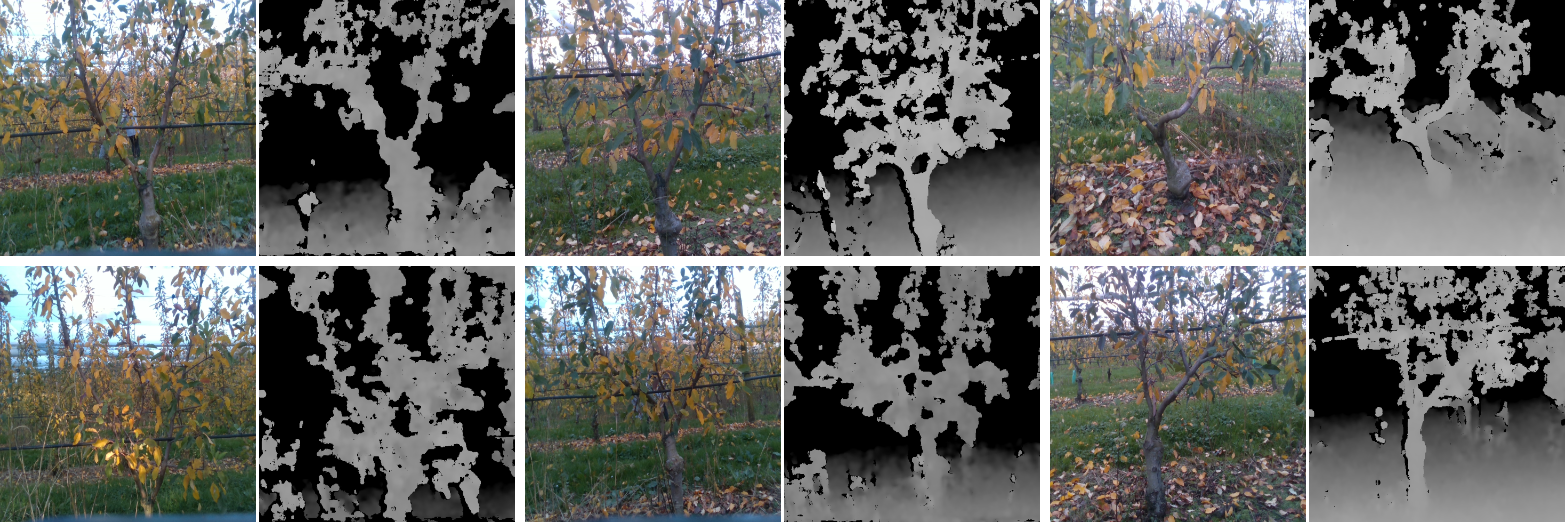}
\caption{Examples of collected RGB and depth images from orchard}
\label{fig:collected data}
\end{figure}

The collected RGB and depth images were center cropped into $480$ by $480$, and then resized into $256$ by $256$ (Figure~\ref{fig:collected data}), before being used as inputs into the neural networks. \citet{MAJEED201875} previously used depth images to threshold the RGB image in their work. However, in our dataset it was noted that particularly thin branches were not recognized by the depth camera. Therefore, rather than using depth information for thresholding, we utilize depth image it as an additional input channel to the neural networks. 

\begin{figure}[h]
         \centering
           \subfloat[RGB Image]{%
              \includegraphics[width=0.2\textwidth]{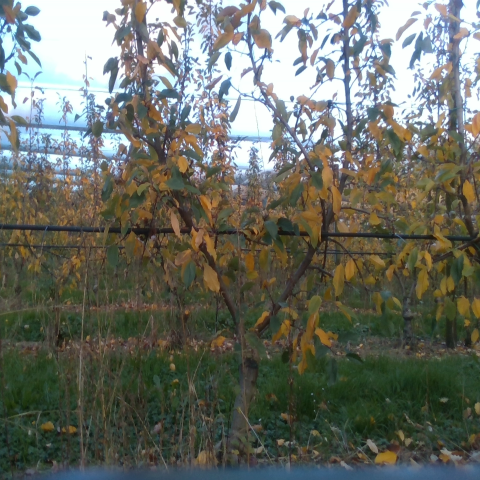}%
              \label{fig:left}%
           }
           \qquad
           \subfloat[Depth Map]{%
              \includegraphics[width=0.2\textwidth]{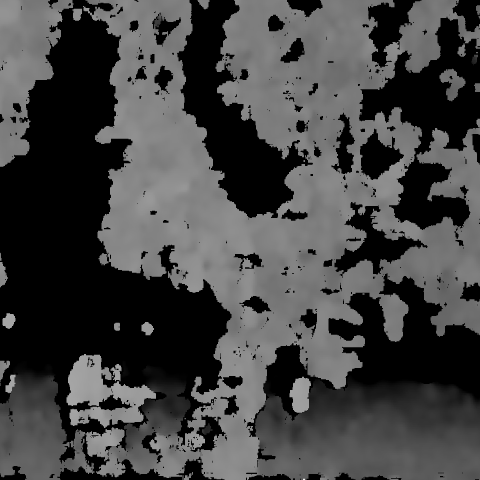}%
              \label{fig:middle}%
           }
           \qquad
           \subfloat[\textit{Branch} Class]{%
              \includegraphics[width=0.2\textwidth]{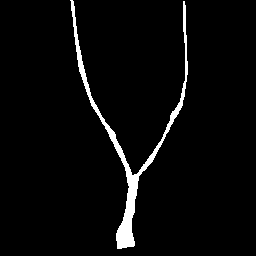}%
              \label{fig:right}%
           }
           \qquad
          \subfloat[\textit{Occluded Branch} Class]{%
              \includegraphics[width=0.2\textwidth]{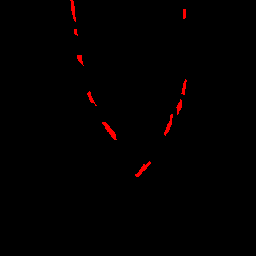}%
              \label{fig:right}%
           }
           \qquad
           \caption{Example of collected data and labels}
           \label{fig:labelexample}
    \end{figure}

The dataset consisted of $521$ hand labeled masks with Adobe Photoshop, seperated into two classes, \textit{Branch} and \textit{Non-Branch}. $50$ images were utilized for the validation set, whilst $471$ were utilized in training. Furthermore, the validation set was additionally labeled with a third class \textit{Occluded Branch} to provide further analysis on the performance of the network. Examples of the labels are shown in Figure~\ref{fig:labelexample}.

\subsection{Segmentation Models}

All semantic segmentation models were trained and tested in Python 3, Tensorflow V 2.3 deep learning framework, using a Nvidia GTX 1060 and CUDA 10.1. They used manually labeled boolean tree skeleton images as the training targets and RGBD (RGB image + Depth map) images as inputs. The models were trained with a batch size of 8 and evaluated after 140 epochs, when all models observed convergence. To update the network weights we utilize the Adam optimizer in it's default configuration on Tensorflow.   All models used Weighted Dice Loss (WDL) unless otherwise specified.

\subsubsection{Pix2Pix}

\begin{figure}[h]
\centering
\includegraphics[width=0.9\textwidth]{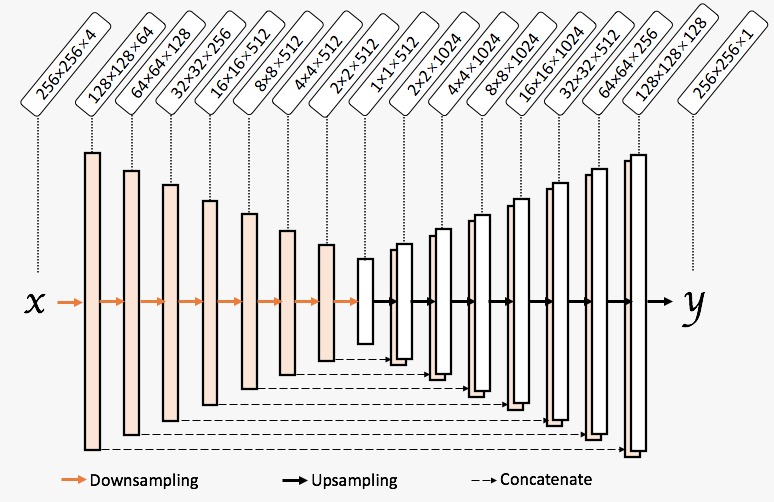}
\caption{The generator network architecture of Pix2Pix}
\label{fig:pix2pix}
\end{figure}

Pix2Pix~\citep{pix2pix} is a conditional GAN that was developed for image transfer and it performs well in mapping pixels to pixels, for example converting maps to aerial photos and Black and White images to color images. The Pix2Pix model used in this project preserved the structure and most of the hyper-parameters from the original paper. The architecture of Pix2Pix's generator network is shown in Figure~\ref{fig:pix2pix}. The generator is a modified U-Net and there are $8$ downsampling blocks in the encoder, with structure of Conv-Batchnorm-LeakyReLu, and another $8$ upsampling blocks in the decoder with structure ConvTranspose-Batchnorm-Dropout-ReLu (Dropout applied to the first $3$ blocks). Skip connections were used to transfer low level feature maps from encoder to the corresponding decoder, which helped to track back the pixel locations in deeper layers. As the output of our model was a binary mask we used a sigmoid activation function for the final.

The discriminator of Pix2Pix is a Patch GAN with $3$ downsampling blocks followed by $2$ convolutions with a stride of $1$. The goal of the discriminator is to distinguish generations from ground truth given the RGBD image. 
The objective of the generator network is:

\begin{equation}
    G^{*} = \arg \min\limits_{{G}} \max\limits_{{D}} \mathcal{L}_{cGAN}(G,D) + \lambda \mathcal{L}_{L1}(G)
    \label{equ:p2ploss}
\end{equation}
where:\\
$G^{*}$ refers the loss function of generator, $G$ represents generator and $D$ represents the discriminator. $\mathcal{L}_{cGAN}$ is the loss function provided by discriminator, $\mathcal{L}_{L1}$ is the mean absolute error between the generator's prediction and the ground truth and $\lambda$ is a hyper parameter to adjust the weighting of the two loss functions.

This function was designed such that it would both fool the discriminator and be near to the ground truth. By minimizing $\mathcal{L}_{L1}$, the objective function encourages the network to be close to the ground truth. Furthermore, \citet{pix2pix} showed that L1 distance encourages less blurring than L2 distance.

$\mathcal{L}_{cGAN}$ is provided by the discriminator, where the network will aim to minimize this loss, which correlates to fooling the discriminator. We set $\lambda=100$ as suggested in \citet{pix2pix}.

In order to best explore both this model's ability to segment occluded image as well as the effect of a GAN training loop, we compare the Pix2Pix generator network with and without the discriminator network in Section~\ref{subsec:p2ploss}. To differentiate the modified Pix2Pix model and the original Pix2Pix model, the original Pix2Pix with discriminator network has been named Pix2Pix GAN and the modified Pix2Pix model with only the generator network has been named Pix2Pix Generator in this work.

To explore the influences of loss function of Pix2Pix models, both L1 loss and WDL function was used to train Pix2Pix Generator as a supervised learning model. The results of Pix2Pix Generator with different loss functions are compared with the results of Pix2Pix GAN in Section~\ref{subsec:p2ploss}. 

\begin{figure}[h!]
\centering
\includegraphics[width=0.7\textwidth]{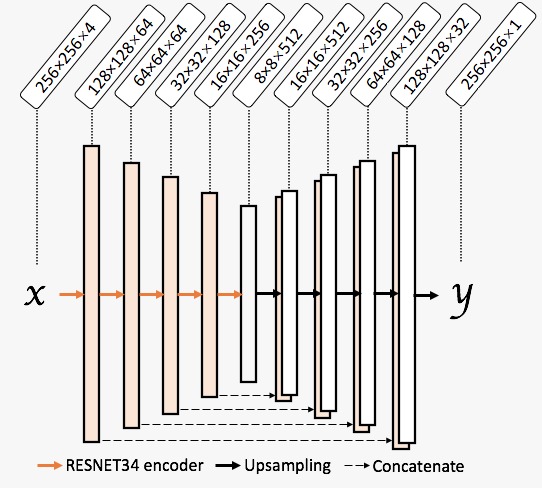}
\caption{Architecture of U-Net}
\label{fig:unet}
\end{figure}

\subsubsection{U-Net}

U-Net~\citep{unet} is a convolutional network for precise semantic segmentation. It is based on fully convolutional network~\citep{fcn} and is efficient at segmenting. U-Net consists of an encoder and a decoder. In order to learn robust features, this project utilized ResNet34~\citep{resnet34} as the encoder, whose intermediate outputs were used by decoder as skip connections, as shown in Figure \ref{fig:unet}.

\subsubsection{DeepLabv3}

DeepLabv3~\citep{deeplab} is a state of the art neural network created by a team at Google. Since it's creation in 2016 it has undergone multiple iterations, with the network utilized in this paper being version $3$. DeepLab utilizes Atrous spatial pyramid pooling (ASPP) in conjunction with the encoder decoder methodology to extract dense feature maps robust to deformation and scalings. ASPP generates multiple scaled versions of it's inputs during training to make the network robust against variable image size and deformation. In order to create the multi-scaled inputs required, DeepLab utilizes Atrous Convolutions. For each Atrous convolution:

\begin{equation}
    output[i] = \sum_{k}input[i + r \times k]w[k]
    \label{eq:atrous}
\end{equation}

\begin{figure}[h!]
\centering
\includegraphics[width=0.7\textwidth]{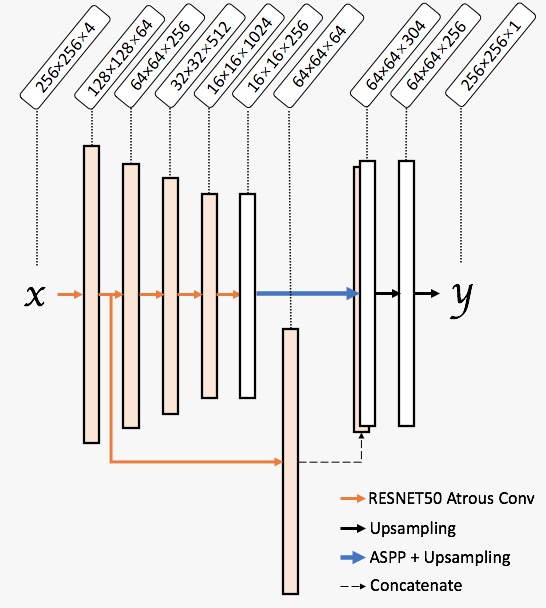}
\caption{Architecture of DeepLabv3}
\label{fig:deeplab}
\end{figure}

\begin{flushleft}where $w$ is a filter and, $i$ and $k$ refers to the pixel locations. These convolutions utilize a rate parameter $r$ which spaces out each pixel that is convolved over in the convolutional layers, allowing variable effective field of view whilst maintaining constant computational complexity and parameter count. In this work, ResNet50~\citep{resnet34} was utilized as the encoder and the decoder used bilinear interpolation. One skip connection was applied to pass low level features, in order to reserve pixel locations in output images, as shown in Figure~\ref{fig:deeplab}.\end{flushleft}

\subsection{Metrics}

Three traditional metrics were used to evaluate the semantic segmentation models' overall performances: Binary accuracy, Mean Intersection-Over-Union (Mean IoU), and Boundary F1 score. Binary accuracy is defined as the percentage of pixels that are correctly classified into either \textit{Branch} class or \textit{Non-branch} class. Mean IoU, also known as Jaccard Index, represents the mean intersection over union of \textit{Branch} and \textit{Non-branch} class. To be able to calculate the thickness of each branch, boundary information is important. Therefore, we use Boundary F1 score to measure the closeness of the predicted boundary and the ground truth boundary.

\begin{equation}
    \textit{Binary Accuracy} = \frac{\textit{TP} + \textit{TN}}{\textit{TP} + \textit{TN} + \textit{FP} + \textit{FN}}
\end{equation}
\begin{equation}
    \textit{IoU} = \frac{\textit{TP}}{\textit{TP} + \textit{FP} + \textit{FN}}
\end{equation}
\begin{equation}
    \textit{Precision} = \frac{\textit{TP}}{\textit{TP}+\textit{FP}}
\end{equation}
\begin{equation}
    \label{eq:recall}
    \textit{Recall} = \frac{\textit{TP}}{\textit{TP} + \textit{FN}}
\end{equation}
\begin{equation}
    \textit{Boundary F1 Score} = \frac{2 * \textit{Precision} * \textit{Recall}}{\textit{Recall} + \textit{Precision}}
\end{equation}

Where TP is true positive, TN is true negative, FP is false positive, and FN is false negative. 

To provide further analysis on the performance under occlusion, we used the Recall metric (Equation~\ref{eq:recall}) on the \textit{Occluded Branch} class and \textit{Branch} class individually.

\subsection{Difficulty Index}

Amodal segmentation is a closely related research topic, however, has not been explored in great detail. \citet{BRECKON2005499} define amodal perception as the "problem of completing partially visible artifacts within a 3D scene". \citet{li2016amodal} provide a comprehensive discussion of the difficulties of amodal segmentation,

During occluded branch segmentation, there are two main factors which contribute to the difficulty of segmentation: Occlusion Amount and Depth Detection Rate. We defined Occlusion Difficulty Index and Depth Difficulty Index correspondingly, and provide analysis in Section~\ref{subsec:occluded} and Section~\ref{subsec:depth} on the $10$ hardest images for each factor.

\subsubsection{Occlusion Difficulty Index}

The more occluded a branch is, the more the network will have to hallucinate and the harder the task is. We defined the occlusion difficulty index as:

\begin{equation}
    \textit{Occlusion Difficulty Index} = \frac{\textit{Occluded Region} \cap \textit{Label} }{\textit{Label}}
\end{equation}
where the Occluded Region corresponds to manual annotations of all occluding leaves, the Label corresponds to manual annotation of \textit{Branch} class (Figure 1c).

\subsubsection{Depth Difficulty Index}

In this project, the depth map was one channel for the input images. This has been utilized in previous research~\citep{ZHANG2018386} to improve the performance of branch segmentation. However, it introduces a potential problem when the depth image contains missing information. This can occur due to sensory factors such as strong light interference or inadequate resolution on extremely subtle objects. As a result, this creates an extra difficulty index representative of real life obstacles that the model may meet. 

From observations of the data set a correlation was found between the thinness of the branch and the inaccuracy of the depth map in identifying the branch. Thus, worse performing re-generations can potentially be attributed to inaccurate depth maps from the sensor that was used. 

The depth difficulty index was defined as below:

\begin{equation}
    \textit{Depth Difficulty Index} = 1-\frac{\textit{Depth detected region} \cap \textit{Label} }{\textit{Label}}
\end{equation}
where the Depth Detected Region corresponds to the areas on the captured depth map that are below the maximum depth threshold of $4$ meters, the Label corresponds to manual annotation of Branch class (Figure 1c).

\section{Results and Discussion}
\label{sec:r+d}

\subsection{Overall performance}

\begin{figure}[h!]
\centering
\includegraphics[width=1\textwidth]{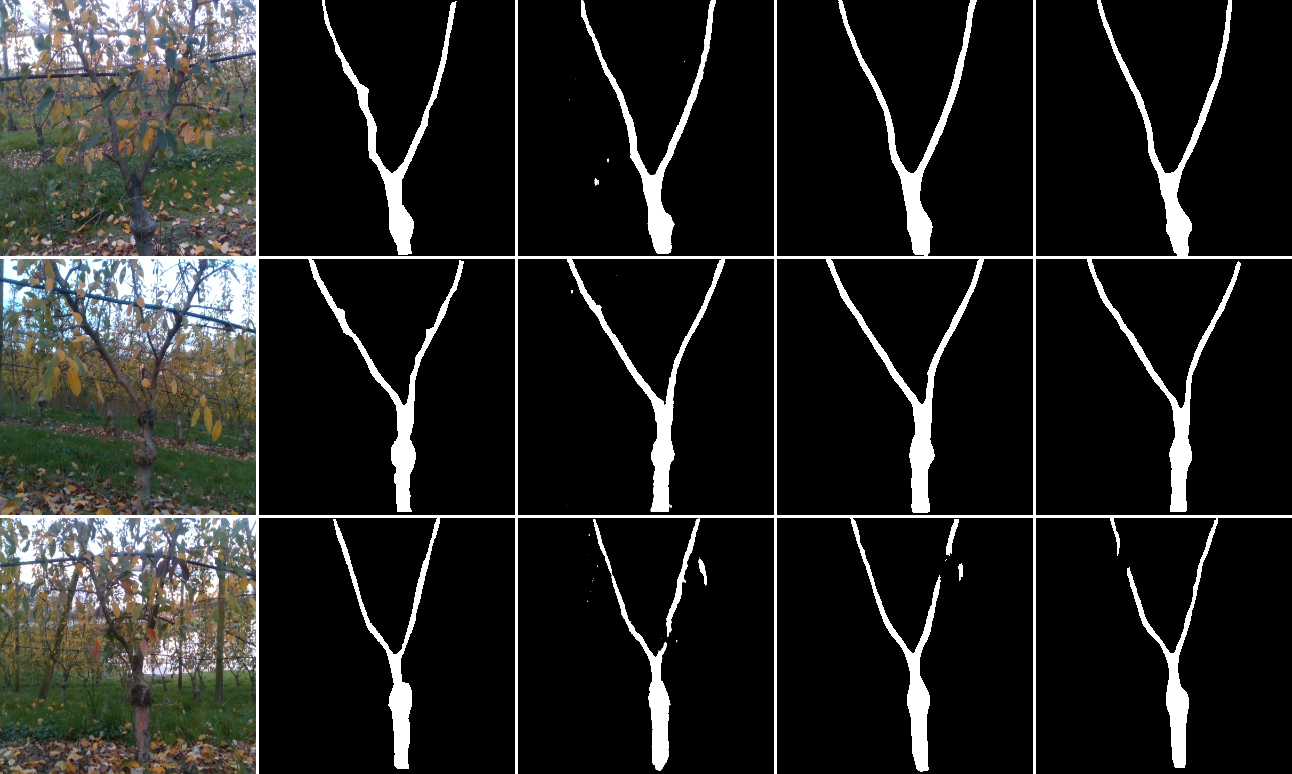}
    \begin{tabular}{*{5}{P{2.0cm}}}
    (a) & (b) & (c) & (d) & (e)
  \end{tabular}
\captionof{figure}{Examples of prediction for the different models: (a) RGB test image; (b) branch ground truth for corresponding RGB image; (c), (d), and (e) are predictions of Pix2Pix Generator, U-Net, and DeepLabv3.}
\label{fig:overallexamples}
\end{figure}

The performance of the models on the validation set are shown in Table~\ref{tab:overall}. DeepLabv3 performed the best in Binary accuracy, Mean IoU, and Boundary F1 score whereas Pix2Pix Generator and U-Net outperformed DeepLabv3 in Occluded Branch Recall. Even has the highest Binary Accuracy, DeepLabv3 got the lowest \textit{Branch} Recall in three models. Pix2Pix Generator's generations were observed to contain more noise, examples of this can be seen in Figure~\ref{fig:overallexamples}. The noise could be a potential contributor to the decrease in Mean IoU and Boundary F1 scores of Pix2Pix Generator when compared to the DeepLabv3 and U-Net. 

Observation of the images in the dataset alludes to the data being heavily unbalanced, potentially contributing to high scores in Binary accuracy (only $5.9\%$ pixels belong to \textit{Branch} class).

\begin{table}[h!]
\begin{center}
    \caption{Pix2Pix, Unet and Deeplab performance on validation set}
    \label{tab:table1}
    \begin{tabular}{lccc}
    \hline%
    \multirow{2}{*}{\textbf{Metrics}} &  \multicolumn{3}{c}{\textbf{Model}}\\ 
    \cline{2%
    -%
    4}%
    & \textbf{P2P Generator} & \textbf{U-Net} &  \textbf{DeepLabv3} \\ 
      \hline
      Binary Accuracy & 97.3\% & 97.8\% & 98.0\%\\
      Mean IoU & 80.2\% & 83.0\% & 83.7\%\\
      Boundary F1 & 81.2\% & 85.3\% & 87.5\%\\
      Branch Recall & 81.8\% & 83.5\% & 79.5\%\\ 
      Non-branch Recall & 98.2\% & 98.6\% & 99.1\%\\ 
      Occluded Branch Recall & 77.6\% & 78.5\% & 74.7\%\\ 
      \hline
    \end{tabular}
\label{tab:overall}
\end{center}
\end{table}

\begin{table}[h!]
\begin{center}
    \caption{Computation time and weights size of Pix2Pix Generator, U-Net and DeepLabv3 on GTX-1060}
    \label{tab:table2}
    \begin{tabular}{lccc}
    \hline
    \textbf{Model} & \textbf{Average Time} & \textbf{Weights Size} \\ 
      \hline
      P2P Generator & 23 ms & 653.0 MB\\
      U-Net & 35 ms & 293.4 MB\\
      DeepLabv3 & 36 ms & 142.1 MB \\
      \hline
    \end{tabular}
\label{tab:computation time}

\end{center}
\end{table}

The comparison of computation time and weights size between three models is shown in Table \ref{tab:computation time}. Note the average time shown here includes both image loading time and inference time. Pix2Pix Generator has the largest weights size, but shortest computation time. U-Net and DeepLabv3 have similar computation time with U-Net doubles the weights size compared with DeepLabv3.

\subsection{Pix2Pix loss function}
\label{subsec:p2ploss}
 
As Pix2Pix was originally used in a GAN structure with adversarial loss, in order to explore the potential of the model, a variety of loss functions were tested. These included the loss functions of the original paper (L1 loss + adversarial loss), L1 loss, and the loss function used with DeepLabv3 and the U-Net (WDL). The results for these functions are shown in Table~\ref{tab:pix2pixloss}.

\begin{table}[h!]
\begin{center}
    \caption{Loss function comparison on Pix2Pix performance on validation set}
    \label{tab:table2}
    \begin{tabular}{lccc}
    \hline
    \multirow{3}{*}{\textbf{Metrics}} &  \multicolumn{3}{c}{\textbf{Model}}\\ 
    \cline{2%
    -%
    4}%
    & \multicolumn{2}{c}{\textbf{P2P Generator}} & \textbf{P2P GAN}\\
    \cline{2%
    -%
    4}%
    & \textbf{Weighted Dice loss} & \textbf{L1 Loss} & \textbf{Hybrid Loss} \\ 
      \hline
      Binary Accuracy & 97.3\% & 97.4\% & 96.5\%\\
      Mean IoU & 80.2\% & 76.7\% & 75.9\%\\
      Boundary F1 & 81.2\% & 80.4\% & 75.3\%\\
      Occluded Branch Recall & 77.6\% & 73.2\% & 65.6\%\\ 
      \hline
    \end{tabular}
\label{tab:pix2pixloss}

\end{center}
\end{table}

It was observed that after removing Discriminator loss, Pix2Pix’s performance was improved in all four metrics. This could be because GAN models commonly suffer from destabilization as they can be highly sensitive to hyperparameters~\citep{arjovsky2017principled}. The hyperparameter configuration was not optimised for this project.
 
 WDL is a geometry based loss function, similar to the definition of Mean IoU and Boundary F1 score. The weight in the loss function was found by calculating the average distribution of \textit{Branch} to \textit{Non-branch} across the entire labeled dataset. Therefore, WDL accommodated for the unbalanced dataset and outperformed L1 loss on these metrics including Occluded Branch Recall. 

\subsection{Evaluation on highest occlusion difficulty scoring images}
\label{subsec:occluded}

\begin{figure}[h!]
\centering
\includegraphics[width=0.7\textwidth]{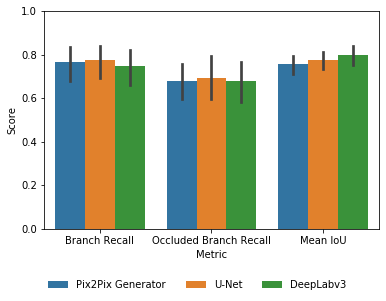}
\caption{Metrics score with P2P Generator, U-Net and DeepLabv3 over the worst 10 occluded images}
\label{fig:occludedhis}
\end{figure}

\begin{figure}[h!]
\centering
\includegraphics[width=1\textwidth]{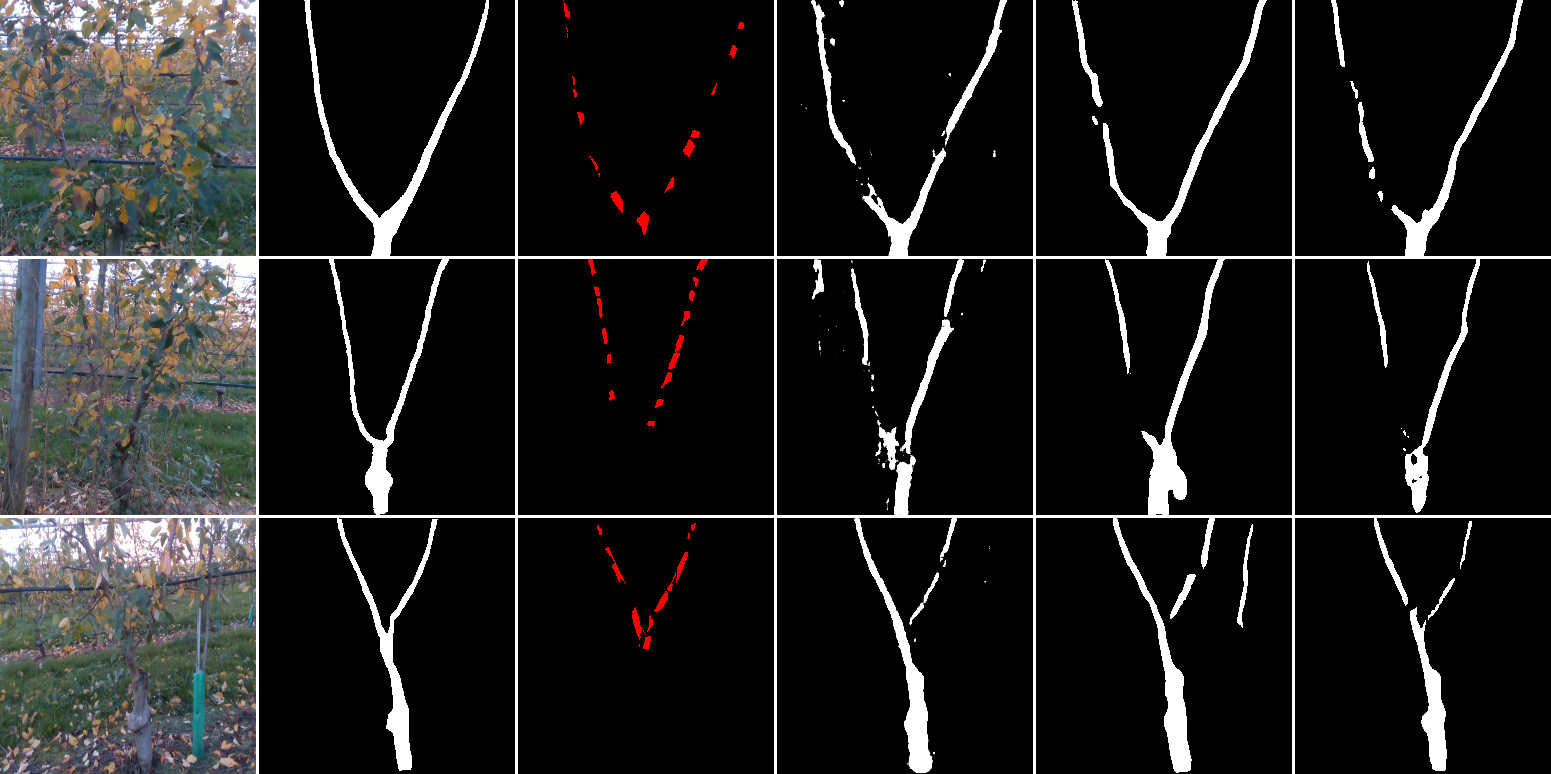}
    \begin{tabular}{*{6}{P{1.6cm}}}
    (a) & (b) & (c) & (d) & (e) & (f)
  \end{tabular}
\caption{Example of prediction for the different models under heavy occlusion: (a) RGB test image; (b) branch ground truth and (c) occluded ground truth for corresponding RGB image; (d), (e), and (f) are predictions of Pix2Pix Generator, U-Net, and DeepLabv3.}
\label{fig:occludedexamples}
\end{figure}

Figure~\ref{fig:occludedhis} shows the scores of Branch Recall, Occluded Branch Recall, and Mean IoU of Pix2Pix Generator, U-Net, and DeepLabv3 under heavily occluded cases. U-Net performed the best at Branch Recall and Occluded Branch Recall. DeepLabv3 outperformed the other two models in Mean IoU. This is potentially because DeepLabv3 has less false positives in the background compared to the other models. This is especially apparent in comparison to the noisy generations of Pix2Pix.

Examples for the three models’ performances under heavily blocked images are shown in Figure~\ref{fig:occludedexamples}. While the Pix2Pix Generator had more noise, some of this noise helped indicate the path of the branch, which is not currently reflected in the metrics. In contrast, U-Net and DeepLabv3 were better at eliminating noise, but left larger gaps along branches in their generations. This highlights the need for more attentive metrics to properly evaluate the models.

In this application the ability to infer occluded branches is considered more valuable than eliminating noise produced from the generator. Noise can be removed through post-processing algorithms. However, without a high degree of accuracy within the occluded areas, key information such as branch endpoints or intersection points may be lost, resulting in incomplete tree structure predictions. 

\subsection{Evaluation on highest depth difficulty scoring images}
\label{subsec:depth}

\begin{figure}[h]
\centering
\includegraphics[width=0.7\textwidth]{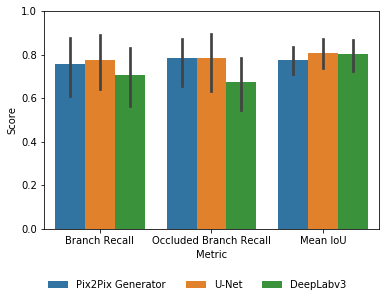}
\caption{Metrics score with P2P Generator, U-Net and DeepLabv3 over the worst 10 depth detection images}
\label{fig:depthhis}
\end{figure}

Figure~\ref{fig:depthhis} summarizes the Branch Recall, Occluded Branch Recall and Mean IoU of Pix2Pix Generator, U-Net, and DeepLabv3 on images with the lowest depth image recognition. U-Net performed the best in all three metrics with Pix2Pix Generator achieved the same score in Occluded Branch Recall. For examples, the first row in Figure~\ref{fig:depthexamples} shows how the left branch failed to be detected by the depth camera resulting in poor predictions of the left branch in all three models. Among these three predictions, however, it can be seen that Pix2Pix Generator detected the most \textit{Branch} pixels from the left branch. An extreme case is shown in the second row of  Figure~\ref{fig:depthexamples} where there is minimal depth information. Whilst all models struggled at detecting the right branch, Pix2Pix Generator was able to detected a few more pixels on the right branch.

\begin{table}[h!t]
  \centering
  \includegraphics[width=1\textwidth]{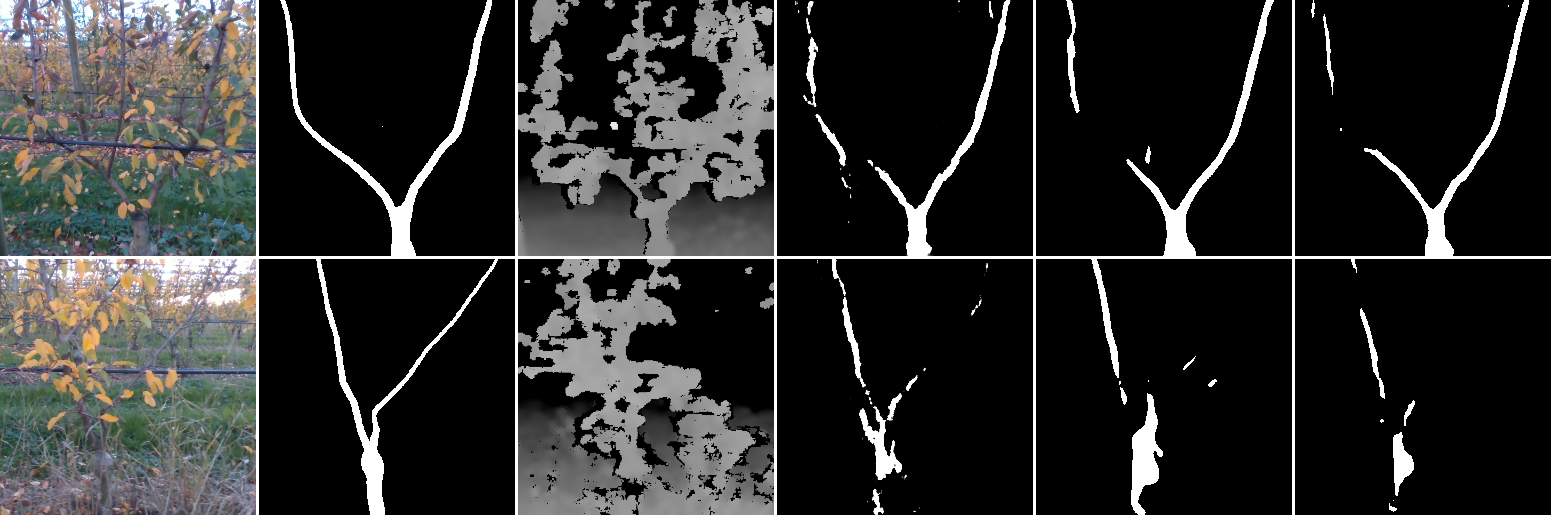}
    \begin{tabular}{*{6}{P{1.6cm}}}
    (a) & (b) & (c) & (d) & (e) & (f)
  \end{tabular}
  \captionof{figure}{Examples of poor depth detection images: (a) RGB and (b) depth images; (c) branch ground truth for corresponding RGBD image; (d), (e), and (f) are predictions of Pix2Pix Generator, U-Net, and DeepLabv3.}
  \label{fig:depthexamples}
\end{table}

In the natural orchard environment, sunlight interference would be unavoidable which makes it important for networks to have a high tolerance to reduced depth information. Both Pix2Pix Generator and U-Net show higher tolerance in this experiment over DeepLabv3.

\section{Conclusion}
\label{sec:conclusion}

 In this work, we compared U-Net~\citep{unet} and DeepLabv3~\citep{deeplab}, two popular semantic segmentation models with the original and modified models of Pix2Pix~\citep{pix2pix} on the segmentation of occluded branches, and found that existing state-of-art segmentation tools are not optimized to recover missing information from occluded regions.

From the results, U-Net is superior to Pix2Pix Generator, Pix2Pix GAN, and DeepLabv3 in the current metrics. The output of Pix2Pix Generator contains more noise than the other models, however, it qualitatively outperforms the other models in detecting occluded branches which provides more information on branch paths. This information can then be utilized in post processing algorithms to further detect tree structure. These results suggests that the traditional metrics for evaluating visible segmentation are not ideal for occluded segmentation. Further research is now required in order to explore new models and modifications to reach the accuracy required for practical implementation in agricultural computer vision systems. 

There are many limitations to the overall performance of this network that could be addressed with other methods. The image dataset of 521 trees will need to be expanded in order to provide for larger generalizability to other trees. Where tree formation follows consistent predictable patterns, such as in our dataset, branches could be repaired with polynomial equation algorithms to recover missing information. Additionally, differentiation of branches and trunks on the labels could potentially provide further information to help the model.

It was noted that whilst the Intel Realsense camera is a highly regarded RGB-D camera, it does suffer in full sun and it was noted that the depth images we took sometimes did not record extremely thin branches. Thus, it could be worth trialing other depth cameras such as the Asus Xtion, or to use LiDAR sensor to capture depth information. Differentiation of the branches and trunk as well as detection of secondary and smaller branches are significant steps to creating a full structure of the entire tree to improve overall understanding of the tree structure.

\bibliography{mybibfile}

\end{document}